\documentclass[a4paper,fleqn]{cas-sc}

\usepackage[authoryear,longnamesfirst]{natbib}
\usepackage{caption}
\usepackage[caption=false,font=normalsize,labelfont=sf,textfont=sf]{subfig}
\usepackage{amsmath,amsfonts}
\usepackage{algorithmic}
\usepackage{algorithm}
\usepackage{array}
\usepackage{textcomp}
\usepackage{stfloats}
\usepackage{url}
\usepackage{verbatim}
\usepackage{graphicx}
\usepackage{color, xcolor}
\usepackage{amssymb}                                \usepackage{float}
\usepackage{bm}
\usepackage{soul}
\soulregister\cite7
\usepackage{balance}

\def\tsc#1{\csdef{#1}{\textsc{\lowercase{#1}}\xspace}}
\tsc{WGM}
\tsc{QE}
\tsc{EP}
\tsc{PMS}
\tsc{BEC}
\tsc{DE}
\begin{document}
\let\WriteBookmarks\relax
\def\floatpagepagefraction{1}
\def\textpagefraction{.001}
\let\printorcid\relax

\shorttitle{Domain Adaptation from Generated Multi-Weather Images for Unsupervised Maritime Object Classification}

\shortauthors{Dan Song et~al.}

\title [mode = title]{Domain Adaptation from Generated Multi-Weather Images for Unsupervised Maritime Object Classification}                      

% % Title footnote 1.
% % eg: \tnotetext[1]{Title footnote text}
% % \tnotetext[<tnote number>]{<tnote text>} 
% \tnotetext[1]{This work was supported by the National Key Research and Development Program of China (2021YFF0704000) and the National Natural Science Foundation of China (U22A2068, U21B2024).}

% \tnotetext[2]{The second title footnote which is a longer text matter
%    to fill through the whole text width and overflow into
%    another line in the footnotes area of the first page.}


\author[1]{Dan Song}

\author[1]{Shumeng Huo}

\author[1]{Wenhui Li}
\cormark[1]
\ead{<liwenhui@tju.edu.cn>} 

\author[1]{Lanjun Wang}

\author[2]{Xue Chao}

\author[1]{An-An Liu}
\cormark[1]
\ead{<anan0422@gmail.com>} 

\address[1]{The School of Electrical and Information Engineering, Tianjin University, China}
\address[2]{Tiandy Technologies Co., Ltd, Tianjin, China}

\cortext[cor1]{Corresponding author}
% \cortext[cor2]{Principal corresponding author}

% % Footnote text
% \fntext[fn1]{This is the first author footnote. but is common to third
% \fntext[fn2]{Another author footnote, this is a very long footnote and
%   it should be a really long footnote. But this footnote is not yet
%   sufficiently long enough to make two lines of footnote text.}

% % For a title note without a number/mark
% \nonumnote{This note has no numbers. In this work we demonstrate $a_b$
%   the formation Y\_1 of a new type of polariton on the interface
%   between a cuprous oxide slab and a polystyrene micro-sphere placed

\begin{abstract}
The classification and recognition of maritime objects are crucial for enhancing maritime safety, monitoring, and intelligent sea environment prediction. 
However, existing unsupervised methods for maritime object classification often struggle with the long-tail data distributions in both object categories and weather conditions. 
In this paper, we construct a dataset named AIMO produced by large-scale generative models with diverse weather conditions and balanced object categories, and collect a dataset named RMO with real-world images where long-tail issue exists.
We propose a novel unsupervised domain adaptation approach that leverages AIMO (source domain) to address the problem of limited labeled data, unbalanced distribution and domain shift in RMO (target domain).
Specifically,we enhance the generalization of source features with the Vision-Language Models (i.e., CLIP) by Self-Knowledge Distillation. 
And we propose a difficulty score for curriculum learning to optimize training process by gradually introducing the generated data.
Experimental results shows that the proposed method significantly improves the classification accuracy, particularly for samples within rare object categories and weather conditions. 
Datasets and codes will be publicly available at \url{https://github.com/honoria0204/AIMO}.
% Datasets and codes will be publicly available after access.
\end{abstract}

% \includegraphics{figs/grabs.pdf}

% % Research highlights
% \begin{highlights}
% \item An effective approach to address the challenge of unsupervised maritime object recognition under various weather conditions and imbalanced samples.
% \item An innovative domain adaptation framework for aligning the generated and real data.
% \item Two datasets, AIMO and RMO, which will contribute to advancing researches in maritime object classification.
% \end{highlights}

% Keywords
\begin{keywords}
Domain Adaptation \sep Maritime Object Classification \sep
\end{keywords}

\maketitle

\section{Introduction}

The classification and recognition of maritime objects is essential for intelligent prediction of the sea environment, with wide applications in maritime traffic supervision, ship rescue and unmanned aerial vehicles (UAVs) cruising. 
Unsupervised maritime object classification based on transfer learning plays a pivotal role in enhancing maritime safety and monitoring by reducing the need for costly labeled data.
In recent years, with the development of deep learning technology, image classification technology based on convolutional neural networks (CNN) and Vision Transformer (ViT) has been gradually applied to maritime object classification, achieving rapid development and breakthroughs.

Unsupervised maritime object classification presents significant challenges in handling the high similarity between ships and complex maritime environments. 
Large-scale annotated datasets are not common because of the difficulty in obtaining maritime objects and the high cost of manual labeling \cite{wang2020generalizing}.
Salem et al. \cite{salem2022transfer} used fine-tuning strategies based on a pre-trained model and achieved good classification results, which also reduced training samples by half.
Zhao et al. \cite{zhao2020maritime} proposed the double transfer method to solve the problem of sample scarcity.
Zhang et al. \cite{zhang2025moment} matched the joint distributions across domains, independent of the availability of source data.
Zhang et al. \cite{DBLP:journals/ipm/ZhangZHM24} considered that the target data are usually highly heterogeneous and designed a Cross-domain Knowledge Collaboration to deal with mixed multiple target domains.
However, the long-tail distribution of object categories, where a few categories dominate and the majority are underrepresented, still poses a critical hurdle.

\begin{figure}
  \centering
  \includegraphics[width=\linewidth]{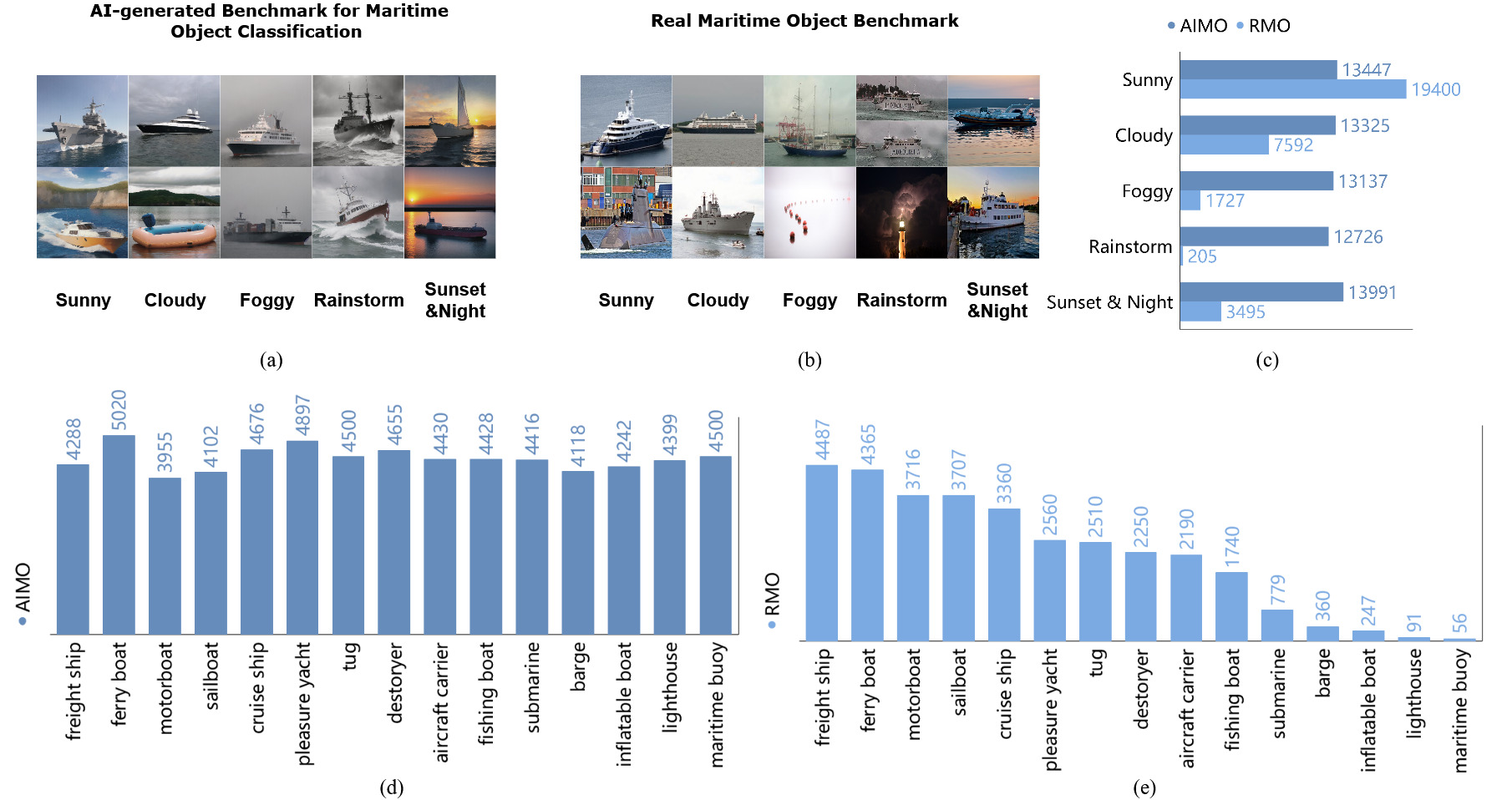}
  \caption{Examples and data statistics of AIMO and RMO.
 (a): Examples of AIMO with multiple weather and illumination conditions. (b): Examples of RMO with multiple weather and illumination conditions. (c): The number of AIMO and RMO with multiple weather and illumination conditions. (d): The number of AIMO with different categories. (e): The number of RMO with different categories.}
  \label{fig:data}
  
\end{figure}

%This classifier imbalance is caused by the long-tail effect, as the maritime object datasets, i.e., MARVEL \cite{gundogdu2017marvel}, ABOships \cite{iancu2021aboships}, Mcships \cite{zheng2020mcships}, and so on, are mostly taken from the real world, where there is a huge difference between the number of objects from different classes.

%Existing domain adaptation techniques, while effective in mitigating domain shift, struggle to maintain performance in the presence of such imbalanced class distributions, leading to suboptimal recognition of less frequent object categories.

The performance of maritime object recognition is also complicated by adverse weather conditions, such as fog, rain, and low visibility. 
DTDNet \cite{liu2022dual} proposed an effective dehazing method to guarantee reliable ship detection under foggy conditions. 
IRDCLNet \cite{sun2022irdclnet} aimed to improve the ship instance segmentation performance on proposed Foggy ShipInsseg dataset. 
However, the methods trained on specific weather phenomena usually fail to generalize across various maritime conditions. 
FREGNet \cite{tian2024fregnet} concentrated on reducing the impact of adverse weather and uneven illumination on supervised object recognition. 
Raza et al. \cite{raza2022simuships} designed SimuShips dataset consisting of maritime objects under multiple weather and time conditions, but this work lacks specific categories for objects, which is only used for obstacle object detection.
In summary, unsupervised maritime object classification under various weather conditions has not been well addressed yet, and new approaches need to be explored.

Due to the shortage of a large label-rich and balanced maritime object data under various weather and illumination conditions, we create AI-generated dataset for Maritime Object classification (AIMO), to compensate for the scarcity of weather conditions in real datasets and address the imbalance in the number of maritime objects across different categories. As shown in Fig. \ref{fig:data},
AIMO is generated by Stable Diffusion, and naturally has a rich set of labels for categories and labels for multiple weather and illumination conditions.
We also construct a real image dataset RMO (Real-world Maritime Object benchmark) by collecting from existing datasets and website. 
Even though advanced diffusion models generate images that are visually indistinguishable from real images, there are still significant differences between their distributions \cite{you2024are}. 
Therfore, we propose a domain adaptation framework in which the generated data serves as the source domain, and the unlabeled real-world data acts as the target domain, facilitating improved classification accuracy under diverse weather conditions in real-world scenarios.

Considering that the generated data may suffer from insufficient diversity and poor generalization, we first enhance the feature representation of the source domain using CLIP \cite{radford2021learning}. Specifically, we design text prompts by combining maritime object categories and weather conditions, and train an image feature extractor to be aligned with the image and text feature spaces of CLIP. Facing the unreliability of the generated source domain and the unlabeled target domain, inspired by Sun et al. \cite{sun2022safe}, we introduce adaptive perturbations to the features to prevent overfitting during training.
Furthermore, we employ a dynamic curriculum learning strategy. We comprehensively consider AIMO's weather and category information, and gradually select samples for training to make model easy to converge.

The contributions of this paper are summarized as follows:
\begin{itemize}
    \item We propose an effective approach to address the challenge of unsupervised maritime object recognition under various weather conditions, effectively handling adverse weather and mitigating the long-tail problem in maritime object classification.
    \item We design an innovative domain adaptation framework for aligning the generated and real data, which overcomes the challenges caused by generated data with the proposed feature generalization enhancement module and curriculum learning strategy.
    % \item We employ a curriculum learning strategy, which dynamically selects the source domain samples to avoid model falling into local optimality.
    \item We construct two datasets, AIMO and RMO, which will contribute to advancing researches in related field.
\end{itemize}

\section{Related Work}

\subsection{AI-generated Datasets}

Currently, one of the most critical challenges in the advancement of Artificial Intelligence (AI) is the lack of high-quality data. 
As a result, there has been considerable attention on image generation technology, although the ongoing debate about the reliability of generated data persists.

Researcher at MIT discovered that, for the person capture task, models trained on generated data (videos with fewer background objects) outperform those trained on real data \cite{zewe2022machine}. 
% Specifically, pre-training with synthetic data has demonstrated superior results compared to pre-training with real data. 
He et al. \cite{DBLP:conf/iclr/HeS0XZTBQ23} identified that larger data volumes and greater data diversity are crucial for achieving better pre-training outcomes with generated data. 
Furthermore, from the perspectives of model architecture and pre-training methodologies, Vision Transformer (ViT)-based models tend to be more effective for pre-training on generated data than convolutional neural networks (CNNs). 
Additionally, self-supervised learning methods have been shown to be more suitable for pre-training with generated data than traditional supervised methods \cite{DBLP:conf/iclr/HeS0XZTBQ23}.

However, recent studies indicate that the use of AI-generated data may lead to model collapse \cite{shumailov2024ai}. 
% This phenomenon occurs when a model forgets long-tail categories or learns a distribution that deviates from the training dataset. 
It is probably due to the generated data not being able to adequately represent the diversity and distribution of real data, which causes the model to gradually overfit these distorted data during training, ultimately compromising its ability to generalize, and leading to model collapse in the worst case. 
% Dohmatob et al. \cite{DBLP:journals/corr/abs-2410-04840} demonstrated that even a 1\% fraction of generated data of the total dataset can significantly disrupt model's performance. 
% Model collapse not only diminishes accuracy, but also undermines the accuracy and reliability of fine-tuned models.
Model collapse not only diminishes accuracy, but also fosters self-reinforcing biases within the model, which further undermines the reliability of models.
% They further pointed out that the capacity of large models may partially offset the negative effects of generated data.
In light of this challenge, our research aims to utilize AI-generated data gradually from easy to difficult samples. 
We leverage the generated dataset, AIMO, for an unsupervised framework based on ViT in maritime object classification. Also, we utilized an \textit{Adaptive Adjustment Mechanism} to address the issues of data overfitting.
To the best of our knowledge, we are the first to overcome the long-tail issue in aspects of object categories and weather conditions for unsupervised maritime object classification by leveraging AI-generated data.

\subsection{Unsupervised Domain Adaptation}

Unsupervised domain adaptation (UDA) seeks to align the source and target domains by learning a feature representation that is invariant across domains.
A series of method for align domains is to minimize the distribution divergence between source and target domains with discrepancy measures, i.e., maximum mean discrepancy (MMD) \cite{kang2019contrastive}, central moment discrepancy (CMD) \cite{zellinger2017central}, correlation alignment (CORAL) \cite{sun2016return} and maximum density divergence (MDD) \cite{li2020maximum}.
Another kind of methods is motivated by adversarial learning, playing a two-player min-max game to learn domain-invariant representations, i.e. DANN \cite{ganin2015unsupervised}, which introduced a domain discriminator to generate domain-invariant features.
Moreover, self-training is used for domain adaptation to generate pseudo-label for the target domain and apply it in the training process \cite{mei2020instance}. 

With the rise of powerful Vision Transformer (ViT) for global feature extraction and modeling, an increasing number of research efforts are focusing on the application of transformers in UDA.
ViT-base and ViT-small are from a pioneering work \cite{DBLP:conf/iclr/DosovitskiyB0WZ21} applying ViT to the task of image classification.
Many works have been proposed on this basis.
Swin \cite{liu2021swin}, \cite{liu2022swin} focused on local self-attention computation and further proposed a large-scale pre-trained model serving as a generic backbone for computer vision.
PMTrans \cite{zhu2023patch} used an intermediate domain to connect the source domain and the target domain, in order to bridge the domain gap, and conducted a min-max cross-entropy game on it.
CDTrans \cite{xu2021cdtrans} proposed cross--attention on source-target image pairs for direct feature alignment.
TVT \cite{yang2023tvt} proposed a multi-head self-attention module to obtain both transferable and discriminative features and combined it with adversarial adaptation.    
% ConMix \cite{kumar2023conmix} Is a Source-free UDA method that predicted pseudo-labels for target domain data through the source model and refines them by using consistency constraints.   

Moreover, considering the strong generalization ability of Vision-Language Models (VLMs), a series of methods focus on the combination of UDA with VLMs. 
It is proved that VLMs are capable of handling multi-modal information (text and image) to further minimizes the domain gap.
Many works have focused on fine-tuning the VLMs.
Cheng et al. \cite{cheng2024disentangled} subsequently disentangled text prompts into domain-invariant and domain-specific descriptions.
Li et al. \cite{li2024split} designed a modality separation network to distinctly disentangle VLM's image embeddings into language-associated and vision-associated components.
However, directly fine-tuning these VLMs into downstream tasks may be computationally expensive, due to the large number of parameters \cite{lai2024empowering}.
Prompt tuning is an efficient learning method with freezing both image and text encoders.
Lai et al. \cite{lai2024empowering} proposed a domain-aware pseudo-labeling scheme designed for VLMs to achieve domain disentanglement.
Bai et al. \cite{bai2024prompt} proposed a Prompt-based Distribution Alignment (PDA) method to integrate the domain knowledge into prompt learning.
DAMP \cite{du2024domain} learned prompts mutually with a cross-attention module based on text embeddings and image embeddings extracted from VLM.
DTKI \cite{zhan2026dual} integrated local feature structures from CLIP with inter-class relationships from the source model to guide adaptation.

Different from the above methods, we utilize a self-knowledge distillation method to transfer the knowledge of VLMs in a flexible and stable way.
We distill the knowledge of text embeddings and image embeddings extracted by frozen VLMs into feature extractor to enhance feature representation, which simultaneously align semantic space and visual space by carrying out bidirectional constraints.

\begin{figure}
% \vspace{-1em}
  \centering
  \includegraphics[width=\linewidth]{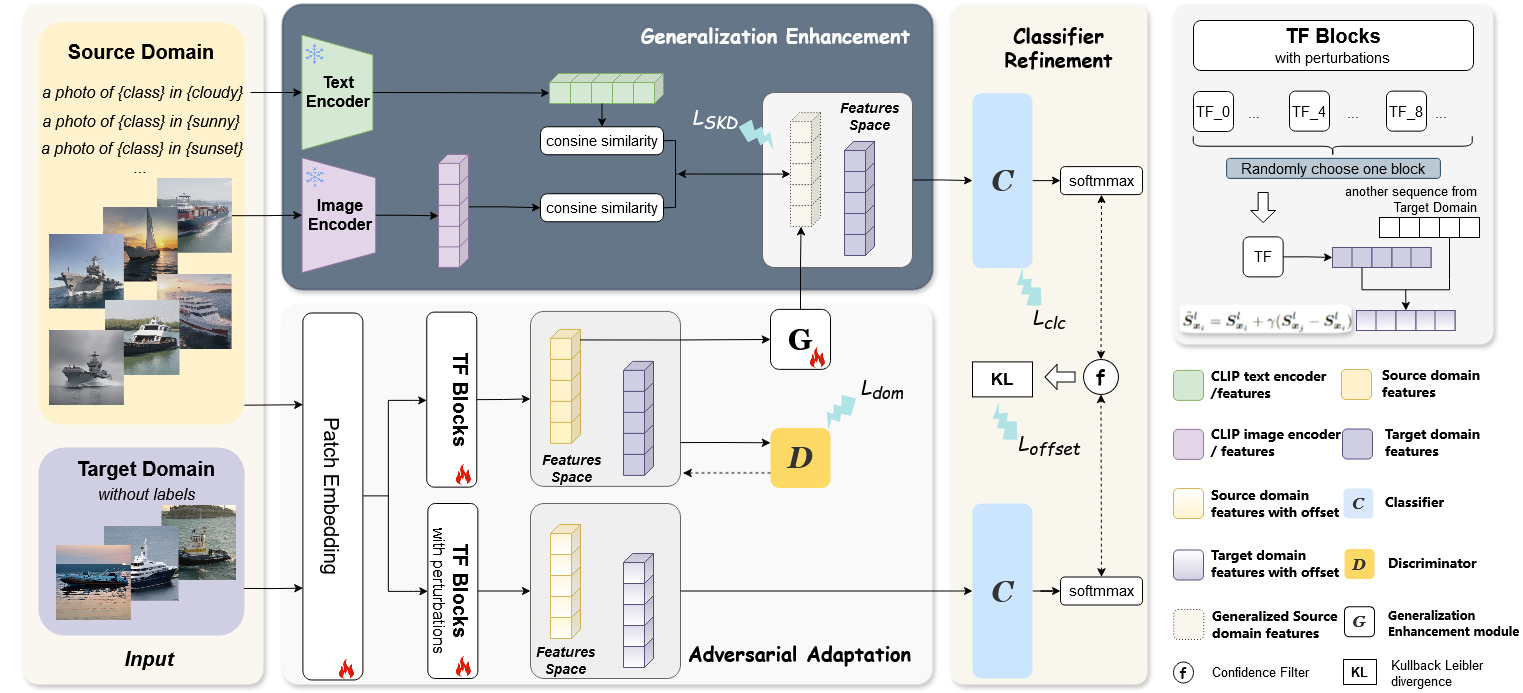}

  \caption{An overview of the proposed method. We use a series of Vision Transformer Blocks as backbone of feature extraction and take labels (classes and weather conditions) and images from the source domain, as well as images from the target domain as inputs. The designed UDA framework for maritime object classification is an adversarial adaptation network, consisting of generalization enhancement for source features, adversarial adaptation from source to target features and classifier refinement with perturbed features from both domains. 
  % The core of AA-CLIP is to narrow the domain gap.
  %Firstly, we propose a \textit{Generalization Layer} for self-distillation of source domain knowledge. The source domain features enhanced through the application of CLIP exhibit superior generalization capabilities.
  % Furthermore, a Adaptive Adjustment Mechanism is utilized to prevent model collapse and data overfitting, as shown on the right.
  %Also, we utilize a novel data augmentation method that introduce a random offset to the input token sequence of a randomly selected transformer block, as shown on the right and refine the model within double branches. 
  }
  \label{fig:method}
% \vspace{-1em}
\end{figure}

\section{Methodology}

%In this section, we define the problem setting and introduce the proposed method. 
We aim to enhance the unsupervised maritime object classification in real-world scenarios by leveraging the generated data.
% with a particular focus the situation of objects with adverse weather conditions and limited sample sizes. 
%Therefore, we propose an adversarial UDA framework based on ViT, as shown in Fig. \ref{fig:method}, utilizing the balanced and label-rich AIMO as source domain. 
The proposed adversarial UDA framework is shown in Fig. \ref{fig:method}, which transfers the knowledge from labeled source domain (balanced generated data) to unlabeled target domain (unbalanced data in reality). The main modules contain generalization enhancement, adversarial adaptation and classifier refinement.

Specifically, the \textit{Generalization Enhancement} aligns the features from Transformer Blocks with the text-encoder and image-encoder features of CLIP for source domain, which could be regarded as a Self-Knowledge Distillation process.
The inputs of CLIP are the source domain data with rich information of category and multiple weather conditions.
We measure the similarity between text embeddings, image embeddings and source domain features to constrain this process.
Inspired by Sun et al. \cite{sun2022safe}, \textit{Classifier refinement} improves the prediction accuracy by enhancing the robustness of feature representation. Random offsets are added into the input token sequence of a randomly chosen transformer block, which is performed on images from both source and target domains to acquire perturbed features.
Then we employ the Kullback-Leibler (KL) divergence to quantify the discrepancy of the corresponding predicted class probabilities of original and perturbed features, and carry out random offset refinement.
Furthermore, we use an \textit{Adaptive Adjustment Mechanism} to carry out the training process step by step by scoring the source domain data.

\subsection{Preliminary}
\label{sec:3.1}
We consider the image classification task in UDA, where a labeled source domain $\mathcal{D}_{s}={\{(\bm{x}^{s}_{i}, \bm{y}^{s}_{i})\}}^{n_s}_{i}$ with $n_s$ examples
and an unlabeled target domain $\mathcal{D}_{t}={\{\bm{x}^{t}_{i}\}}^{n_t}_{i}$ with $n_t$ examples are given. 
Note that the two domains share the same label space $\mathbb{R}^K$, where K is the number of classes.

The basic adversarial UDA framework is composed of three parts: feature extractor \textbf{\textit{F}}, consisting of a Patch Embedding layer and a series of  Transformer Blocks, classifier \textbf{\textit{C}}, and discriminator \textbf{\textit{D}}.  
% For each image, the Patch Embedding layer transforms it into a token sequence including a special class token and image tokens.
% Then the sequence is refined with a series of Transformer Blocks. 
Given an input $\bm{x}$ from either source domain or target domain, $\bm{S} = \textbf{\textit{F}}(\bm{x})$ is referred to a token sequence encoded by \textbf{\textit{F}}, which is passed to \textbf{\textit{D}} directly.
Our novel adversarial UDA framework contains a \textit{Generalization Enhancement} module \textbf{\textit{G}} for enhancing the source domain feature.
Therefore, the source domain feature $\bm{S}_{\bm{x}^{s}}$ is passed to \textbf{\textit{G}}, and then $\bm{S}^{\textbf{\textit{G}}}_{\bm{x}^{s}}=\textbf{\textit{G}}(\bm{S}_{\bm{x}^{s}})$ is passed to \textbf{\textit{C}}.
The \textbf{\textit{C}} predict $\bm{S}^{\textbf{\textit{G}}}_{\bm{x}^{s}}$ into class probabilities $\bm{p}=\textbf{\textit{C}}(\bm{S}^{\textbf{\textit{G}}}_{\bm{x}^{s}}) \in \mathcal{R}^K$. 
Therefore, we calculate the standard cross-entropy loss on source domain data as classification loss.

\begin{equation}
\label{loss_ce}
\mathcal{L}_{ce} = -\mathbb{E}_{(\bm{x}, \bm{y})\in \mathcal{D}_{s}} \sum_{k=1}^{K} \mathbb{1}_{k=y}\log{\bm{p}_k}.
\end{equation}

Considering the unbalanced samples, we further design a focal loss based on the cross-entropy loss.

\begin{equation}
\label{loss_clc}
\mathcal{L}_{focal} = -\mathbb{E}_{(\bm{x}, \bm{y})\in \mathcal{D}_{s}} \sum_{k=1}^{K} \mathbb{1}_{k=y}(1-\bm{p}_k)^{\tau}\log{\bm{p}_k},
\end{equation}
when $\tau=0$, focal loss degenerates to cross-entropy loss.
And the larger t is, the greater the weight of the inaccurate samples in the loss. 
We will detail the choice of the value of $\tau=0$ in Sec \ref{sec:3.4}.

Meanwhile, \textbf{\textit{D}} predicts $\bm{S}$ into domain logits $\bm{q}=\textbf{\textit{D}}(\bm{s}) \in \mathcal{R}^2$, which is actually a binary domain discrimination to learn domain-invariant feature.
We define the domain adversarial loss $\mathcal{L_{D}}$ as

\begin{equation}
\label{loss_dom}
\mathcal{L}_{dom} = -\mathbb{E}_{\bm{x}\in \mathcal{D}_{s}} \log{\bm{q}_s} - \mathbb{E}_{\bm{x}\in \mathcal{D}_{t}} \log{\bm{q}_t},
\end{equation}
where $\bm{q}_s$ and $\bm{q}_t$ represent source and target domain, respectively.
To achieve domain alignment, a domain-invariant feature encoder is needed to confuse the domain discriminator. Therefore, we adversarially train \textbf{\textit{D}} to minimize $\mathcal{L}_{dom}$ and \textbf{\textit{F}} to maximize $\mathcal{L}_{dom}$, which is implemented by reversing the gradients flowing from \textbf{\textit{D}} to \textbf{\textit{F}}.

The objective of UDA is formulated as
\begin{equation}
\label{loss_all}
\min_{\textbf{\textit{F}}, \textbf{\textit{C}}} \max_{\textbf{\textit{D}}} \mathcal{L}=\mathcal{L}_{focal}-\mathcal{L}_{dom}+\alpha \mathcal{L}_{SKD}+\beta \mathcal{L}_{offset},
\end{equation}
where $\alpha$ and $\beta$ are trade-off parameters introduced in Sec \ref{sec:3.4}. $\mathcal{L}_{SKD}$ represents the Self-Knowledge Distillation (SKD) loss introduced in Sec \ref{sec:3.2} and $\mathcal{L}_{offset}$ represents the random offset refinement loss on both source and target domain introduced in Sec \ref{sec:3.3}.

\subsection{Generalization Enhancement Based on CLIP}
\label{sec:3.2}
VLMs exhibit exceptional performance across various distributions, attributed to the vast diversity of distributions seen during their training \cite{radford2021learning}, which leverage contrastive pre-training methods \cite{zhai2022lit} for learning the correlation between the embeddings of matching image-text pairs.
For instance, CLIP \cite{radford2021learning} are trained jointly on 400 million image-text pairs, yielding remarkable generalization across different data distributions.
Despite the considerable zero-shot generalization of VLMs in classification tasks, effectively adapting them to downstream tasks remains a great challenge.
It is expected to leverage the knowledge of VLM in a portable way.
Knowledge distillation not only avoids large-scale parameter tuning of VLMs, but also helps to realize lightweight reasoning.
For UDA task, it is an effective strategy that enhances cross-domain representation across the open semantic and visual space of VLMs, in order to bridge the domain gap.
Compared to a single label, VLMs have more descriptive information, which is beneficial of making full use of the AIMO's rich label information as semantic consistency constraint.
Therefore, we propose a \textit{Generalization Enhancement} module to efficiently distill the abundant knowledge from CLIP into the adversarial UDA network.
We distill the knowledge of text embeddings and image embeddings extracted by frozen VLMs into feature extractor of ViT to enhance feature representations, by carrying out a two-branch constraint to simultaneously align backbone features with semantic information and visual information from VLMs.

Specifically, we perform a Self-Knowledge Distillation on the source domain data based on CLIP.
The feature invariance is constructed in the source domain by Self-Knowledge Distillation, which makes the model more robust to the distribution migration in the target domain.
Firstly, We utilize the frozen text encoder and image encoder of VLMs to extract embedded features, which update the source domain image representations as a "teacher signal". 
And then the similarity between text embeddings, image embeddings and backbone features is calculated respectively, as constraints to supervise the training process, in order to realize a double alignment of semantic space and visual space.
Given labels for category and weather conditions, as well as images from the source domain as inputs to CLIP, the source domain features of Transformer Blocks are aligned with the text-encoder and image-encoder features of CLIP in the \textit{Generalization Enhancement} module.
As proper text prompt allow for better control of invariance in text embeddings \cite{addepalli2024leveraging}, we use \textit{"A photo of a \{class\} in \{domain\}"} to represent the semantic embedding.
\textit{"\{domain\}"} is referred to the weather and illumination condition of the source image.
Also, we compared the classification results under various prompt templates in Sec \ref{sec:4.5}.

We denote the CLIP’s text embedding for "\textit{A photo of a \{class\} in \{domain\}"} and image embedding of the source domain input $(\bm{x}^{s}_{i}, \bm{y}^{s}_{i})$ as $\bm{T}_{{y}^{s}_{i}}$ and $\bm{I}_{\bm{x}^{s}_{i}}$.
The generalization enhanced feature corresponding to the input $\bm{x}^{s}_{i}$ are denoted as $\bm{S}^{\textbf{\textit{G}}}_{\bm{x}^{s}_{i}}$. 
The following Self-Knowledge Distillation loss is used to refine the aligned features:
\begin{equation}
\label{loss_SKD}
\mathcal{L}_{SKD} =-\mathbb{E}_{(\bm{x}, \bm{y})\in \mathcal{D}_{s}} \sum_{i=1}^{n_s} \{\cos(\bm{S}^{\textbf{\textit{G}}}_{\bm{x}^{s}_{i}}, \bm{T}_{{y}^{s}_{i}}) +\cos(\bm{S}^{\textbf{\textit{G}}}_{\bm{x}^{s}_{i}}, \bm{I}_{\bm{x}^{s}_{i}})\}.
\end{equation}
where $\cos(\cdot, \cdot)$ denotes the cosine similarity between two vectors, and the objective of $\mathcal{L}_{SKD}$ is to distill the comprehensive image features and semantic representations learned by the image encoder and text encoder of the CLIP.

\subsection{Classifier Refinement}
\label{sec:3.3}

We regularize the latent feature spaces of the transformer backbone through classifier refinement using perturbed source and target domain features to reduce overfitting .
Incorporating a random offset into the data is a prevalent data augmentation technique that is widely employed to alleviate overfitting.
It may be better to perform data augmentation operations at hidden layers rather than directly manipulating the input images \cite{verma2019manifold}.
%ViT possesses distinctive architecture.
The optimal layer for introducing perturbations differs from various tasks.
Perturbing relatively deep layers tends to yield superior performance but increases the risk of model collapse, and randomly choosing one transformer block proves to be more robust compared with perturbing any individual layer from them \cite{sun2022safe}.
Therefore, we select a transformer block at random, and then add a random offset to the input token sequence of the chosen block.

We perform regularization on both the source and target domains.
Given an image $\bm{x}_i$ from either source domain or target domain, $\bm{S}^l_{\bm{x}_i}$ represents its input token sequence of the l-th transformer block.
$\bm{S}^l_{\bm{x}_i}$ can be regarded as a latent representation of $\bm{x}_i$ within a hidden space.
And then we utilize the token sequence $\bm{S}^l_{\bm{x}_j}$ of another randomly chosen image $\bm{x}_j$ to add an offset. The perturbed token sequence of $\tilde{\bm{S}}^l_{\bm{x}_i}$ is obtained as
\begin{equation}
\label{random offset}
\tilde{\bm{S}}^l_{\bm{x}_i}=\bm{S}^l_{\bm{x}_i} + \gamma(\bm{S}^l_{\bm{x}_j}-\bm{S}^l_{\bm{x}_i}),
\end{equation}
where $\gamma$ is a trade-off parameter introduced in Sec \ref{sec:3.4}.
Note that gradients do not back-propagate through the offset, while they pass through $\bm{S}^l_{\bm{x}}$.

We use KL divergence to measure the distance of the predicted probability vectors $\bm{p}$ and $\bm{\tilde{p}}$, corresponding to $\bm{S}^l_{\bm{x}}$ and $\tilde{\bm{S}}^l_{\bm{x}}$, respectively. 
The following equation shows the KL divergence between the perturbed  predicted probability and the original predicted probability. 
\begin{equation}
\label{KL}
\bm{D}_{\text{KL}}(\bm{p}||\bm{\tilde{p}})=\sum_{i=1}\bm{p}(i)\log\frac{\bm{p}(i)}{\bm{\tilde{p}}(i)}.
\end{equation}
$\bm{D}_{\text{KL}}(\bm{\tilde{p}}||\bm{p})$ is the calculated similarly. And it is more robust to combine the two KL divergence together in random offset loss \cite{sun2022safe}.
To achieve reliable predictions, we define a Confidence Filter $\mathcal{F}$ as 
\begin{equation}
\label{confidence_filter}
\mathcal{F}[\mathcal{D};\bm{p}] =\{\bm{x} \in \mathcal{D}| \max(\bm{p})> \kappa\},
\end{equation}
where $\kappa$ is predefined threshold.

The random offset loss is defined as an average on every training batch $\mathcal{B} \subseteq \mathcal{D}_{\bm{s}} \cup \mathcal{D}_{\bm{t}}$, containing images from source domain and target domain.
\begin{equation}
\label{offset_loss}
\begin{split}
\mathcal{L}_{offset} = \mathbb{E}_{\mathcal{B} \subseteq \mathcal{D}_{\bm{s}} \cup \mathcal{D}_{\bm{t}}} \{ \omega \mathbb{E}_{\bm{x} \in \mathcal{F}[\mathcal{B}; \bm{p}]} \bm{D}_{\text{KL}}(\bm{p} || \bm{\tilde{p}})  + (1 - \omega) \mathbb{E}_{\bm{x} \in \mathcal{F}[\mathcal{B}; \bm{\tilde{p}}]} \bm{D}_{\text{KL}}(\bm{\tilde{p}} || \bm{p}) \},
\end{split}
\end{equation}
where $\omega$ is a random variable drawn from a Bernoulli distribution $\mathcal{B}(0.5)$ \cite{sun2022safe}. 
Note that the loss gradient is back-propagated through both the perturbed and the original predicted probabilities.

In addition to the random perturbation, the multimodal information is introduced only for the unperturbed source domain images in Sec \ref{sec:3.2} and the Dropout layer within the classifier head operates in a stochastic manner for both branches, which introduce additional source of discrepancy for the random offset refinement loss.

\subsection{Adaptive Adjustment Mechanism}
\label{sec:3.4}
Domain adaptation using synthetic data remains a challenging task, especially for long-tailed distributed maritime objects across diverse weather conditions and multiple categories, where substantial variations in sample difficulty exist.  
Accordingly, we propose an \textit{Adaptive Adjustment Mechanism} based on curriculum learning and adversarial training, in order to avoid model falling into local optimal, reduce data overfitting and avoid possible model collapse.
%In the process of optimizing the training objective, we observe that arranging samples in a meaningful order leads to better performances.
By imitating the "step by step" learning process of human, the curriculum learning gradually guides the adversarial training to realize the alignment of source domain and target domain.
We make models quickly establish basic representations by firstly feeding easier samples, and then gradually introducing the harder ones.
Traditional curriculum learning relied on prior knowledge (usually using fixed values with experience) to evaluate sample difficulty, which leads to lower effectiveness in the later period of training \cite{wang2021survey}.
Recent method \cite{DBLP:journals/corr/abs-2502-04628} tried using losses as dynamic difficulty evaluation.
In this paper, we propose the difficulty score for our task as a combination of prior score and dynamic loss (Eq. \ref{difficulty_score}) to select source domain samples (i.e., the generated data) for training. 
%we design a comprehensive score to evaluate the sample hardness, which introduce an adaptive weight $\phi$ to balance the prior quality and model loss.
%The higher prior quality (clearer object and simpler background of image) and the less model loss leads to the higher final comprehensive score.

%\textcolor{red}{
Firstly, we design a prior score as the Image Quality Assessment (IQA) score weighted by weather condition.
Specifically, we utilize CLIP IQA \cite{pyiqa} for the prior IQA analysis of the source domain images.
According to different weather conditions, the pre-defined weights are:\{\textit{sunny: 5, sunset \& night: 4.5, cloudy: 4, foggy:3.5, rainstorm: 3}\}. Subsequently, higher image quality and better weather condition contribute to a higher prior score.
Secondly, the rest part of difficulty score is based on the standard cross-entropy loss (Eq. \ref{loss_ce}, where more certain category prediction leads to higher dynamic score.
The difficulty score is formulated as follows:
\begin{equation}
\label{difficulty_score}
{difficulty}\_{score} = (1-\phi) {prior}\_{score} + \phi (1 - \mathcal{L}_{ce}),
\end{equation}
where the initial value of the adaptive weight $\phi$ is 0 and gradually increases to 1 at the end of the training in a linear function.
%}

Therefore, the selected source domain data for the t-th epoch is:
\begin{equation}
\label{source domain data}
{A}_{t} = \operatorname{argmax}_{\hat{A} : |\hat{A}| \geq \lambda(t) \cdot |A|} {difficulty}\_{score},
\end{equation}
where $A$ denotes the source domain data, the $|A|$ and $|\hat{A}|$ represent the size of $A$ and $\hat{A}$. $\lambda(t)$ is used to schedule the proportion of source training samples at the t-th epoch, which is defined as:
\begin{equation}
\label{lambda}
% \lambda(t) = \lambda_{0} + (1 - \lambda_{0}) [1 - \exp\left(\frac{-kt}{T}\right)],
\lambda(t) = \lambda_{0} + (1 - \lambda_{0}) (1 - e^{-kt/T}),
\end{equation}
where $\lambda_{0}$ is the initial proportion of source training samples, which is set to 0.5, and $T$ is the total number of epochs. Note that $k$ is set to 2 for the amount of data increasing rapidly in the early stage of training and slowly in the late stage of training, in order to avoid making it difficult to generalize the model to difficult samples due to long-term training of high-score samples.
Also, the $\tau = 5t/T$ in the focal loss varies linearly with $t$ in the range [0, 5], which means that the model loss will gradually pay more attention to difficult samples as the training goes on.

Furthermore, in the classifier refinement strategy, the selection of appropriate values for the perturbation scalar $\gamma$ and the random offset refinement loss weight $\beta$ is crucial.  
Excessively large perturbations result in a collapse of the predicted class distribution, whereas small perturbations may fail to fully leverage their potential benefits.  
Given that the target domain is entirely unlabeled and domain adaptation tasks vary significantly even within the same dataset, it is desirable to adjust these parameters adaptively.
At the initial stage of training, the model is not stable, and too much perturbation may lead to convergence difficulties. 
With the increase of the number of iterations, the proportion of disturbance should be appropriately increased to reduce overfitting.
Inspired by SSRT \cite{sun2022safe}, we plan to introduce random perturbations dynamically, by setting an an adaptive scalar.
% And Sun et al. \cite{sun2022safe} observed that model collapse is invariably accompanied by a concurrent reduction in the diversity of model predictions.
% The subsequent issue pertains to the selection of an appropriate diversity measure and the detection of diversity reductions.   
% Sun et al. \cite{sun2022safe} found that the number of unique model predicted labels within each target training batch $\mathcal{B}_{\bm{t}}$ serves as an effective metric. 
% Therefore, to monitor the model's performance, we measure the unique model predicted labels on the target domain data. 
% Once a decrease in diversity metric occurs, which means that a model collapse has been detected, the learning configuration is reset and the model is reverted to a previously attained state.

% Firstly, the training process is divided into contiguous intervals with a fixed period N.
% At the termination of each interval, a model snapshot is captured and stored.
Specifically, the adaptive scalar $\mu \in [0, 1]$ is defined as
\begin{equation}
\label{mu}
\mu(n) = 
\begin{cases} 
\sin\left(\frac{\pi}{2N}{n}\right) & \text{if } n < N \\
1.0 & \text{otherwise}
\end{cases},
\end{equation}
where n points to the current training iteration, and N is a fixed period, setting to 1000.
Therefore, r gradually goes up to 1.0 after N steps.
% Also, we split the interval into sub-intervals.
% At the end of each sub-interval, the diversity of model predictions is checked to find whether the average diversity value drops across each sub-interval.
% If the diversity decrease is detected, $n_{\mu}$ is reset to current training step $n$, and the model is restored to the last snapshot. To avoid oscillation between
% collapse and restoration, $N_{\mu}$ is doubled if the last restoration occurs within $N_{\mu}$ steps.
% The specific process is shown in Fig .\ref{fig:collapse}.
The adaptive scalar $\mu$ is adopted to modulate $\gamma$ and $\beta$, i.e., $\gamma_{\mu}=\mu \gamma$ and $\beta_{\mu}=\mu \beta$, which builds a bridge between the adaptive adjustment and random offset refinement, making our proposed UDA framework more resilient to avoid data overfitting.

\section{Experiments}

\subsection{Dataset Construction}
\label{sec:4.1}
%We conduct experiments on our proposed AIMO and RMO.
%To address the long-tail issues in both categories of maritime objects and types of weather conditions, 
To overcome the issues of single weather condition and long-tail distribution of maritime objects in existing datasets, we construct an AI-generated dataset, AIMO, and a real-world dataset, RMO with labels of diverse weather conditions and object categories.

AIMO consists of 66,626 images with category labels and multiple weather and illumination conditions.
Taking advantage of the strong generation capabilities of Stable Diffusion-WebUI\footnote{\url{https://github.com/AUTOMATIC1111/stable-diffusion-webui.git}}, we create most of the maritime objects using simple text prompts.
In order to make the generated data closer to the real data, we utilize techniques such as LoRA, image embeddings \cite{gal2022image} and hypernetworks \cite{ha2016hypernetworks} to assist image generation.
For instance, the generated results of warship series, e.g., aircraft carriers, are commonly not satisfactory, so we utilize a LoRA model \footnote{\url{https://www.seaart.ai/models/detail/f8078759a92ae10610a92f8683f893da}} for better generation quality.
The use ratio of LoRA is manually adjusted, according to the generation quality of different categories.
In addition, for some categories that are beyond the coverage of the generated model, e.g., barges, tugs and so on, we select a small amount of real maritime objects images for specific categories to make image embeddings by extracting features. 
The produced image embeddings serve as a prior pseudo-prompt in the generation process, in order to enhance the authenticity and diversity of the generated samples.
Considering the complexity of the sea surface environment, we train a hypernetwork on WSODD \cite{zhou2021image} to achieve sea surface environment style patches, which make the generated data cover a wider range of weather and illumination conditions.
As a small neural network, hypernetworks dynamically adjust the weight parameters of the main model (i.e., Stable Diffusion).

RMO consists of 32,418 images collected from multiple maritime object datasets and websites. 
The main sources of RMO are: Marvel \cite{gundogdu2017marvel}, Mcships \cite{zheng2020mcships}, ABOships \cite{iancu2021aboships}, MAR-ships \cite{bloisi2015argos}, CIB-ships \cite{meng2022fine}, WSODD \cite{zhou2021image}, DVTR \cite{zhang2024viewpoint}, Ships Image Dataset\footnote{\url{https://www.kaggle.com/datasets/vinayakshanawad/ships-dataset}}, Boat types recognition\footnote{\url{https://www.kaggle.com/datasets/clorichel/boat-types-recognition}}, Buoyandboat Computer Vision Project\footnote{\url{https://universe.roboflow.com/univeristy-of-southeastern-norway/buoyandboat}}. 
Then we filter these data to form the RMO.
Specifically, we exclude images that lack the object, as well as those depicting the object in sketch form.
And for images containing multiple objects, we crop out each object separately.

The proposed AIMO involves five kinds of weather and time conditions, namely sunny, cloudy, foggy, rainstorm, and sunset \& night.
AIMO shares the same label space as RMO, covering 15 categories: aircraft carrier, barge, cruise ship, destoryer, ferry boat, fishing boat, freight ship, inflatable boat, lighthouse, maritime buoy, motorboat, pleasure boat, sailboat, submarine and tug. 
In order to better verify the effect of AIMO in real scenarios with different weather and illumination conditions, we manually divided the RMO into the above conditions. Note that the annotations are just used for validation where the labels are not accessible during training.

For the UDA task, AIMO and RMO are used as source domain and target domain, respectively. 
We use the average classification accuracy on RMO as the evaluation metric.

\subsection{Training Details}
\label{sec:4.2}
We initialize ViT-B/16 \cite{DBLP:conf/iclr/DosovitskiyB0WZ21} with ImageNet \cite{russakovsky2015imagenet}.
We use the SGD optimizer \cite{le2011optimization} with a constant learning rate of 0.002 for all our experiments and we set batch size to 16.
Our backbone network contains 12 layers of ViT blocks, and we choose blocks 0, 4, and 8 as alternate layers. 
For features of each batch, a perturbation layer is randomly selected from the alternate layers, and the offsets are respectively added for the source domain and the target domain features in this layer.
We set the hyper parameters $\alpha=0.3$, $\beta=1$, $\gamma=0.2$, and $\kappa=0.4$, according to the discussion in Sec \ref{sec:4.6}.

\subsection{Effectiveness of AIMO}
\label{sec:4.3}
In this section, we evaluate the image quality of AIMO, and design experiments to verify AIMO has the ability to improve the accuracy of maritime object classification in complex environments, and solve the classifier imbalance caused by the long-tail effect.

Firstly, we selected three no-reference image quality assessment metrics \cite{pyiqa}, ENTROPY, NIQE, and CLIPIQA to evaluate the quality of AIMO . 
ENTROPY is used to measure the complexity and information of images.
NIQE is used to assess the natural quality of an image.
CLIPIQA is used to evaluate image quality by using CLIP.
Higher values of these metrics indicate better image quality.
Also, we calculated the FID \cite{pyiqa} metric between AIMO and RMO, which characterizes the difference in feature distribution. The results are shown in Table \ref{tab:IQA}.
ENTROPY over 7 and CLIPIQA close to 0.8 indicate that AIMO images have rich details, which contain valuable information to transfer. 
However, the lower NIQE and higher FID indicate a clear difference between AIMO and natural maritime object images, which requires effective domain adaptation.
%In conclusion, our proposed AIMO is acceptable in quality.

\begin{figure}
% \vspace{-1em}
  \centering
  \includegraphics[width=\linewidth]{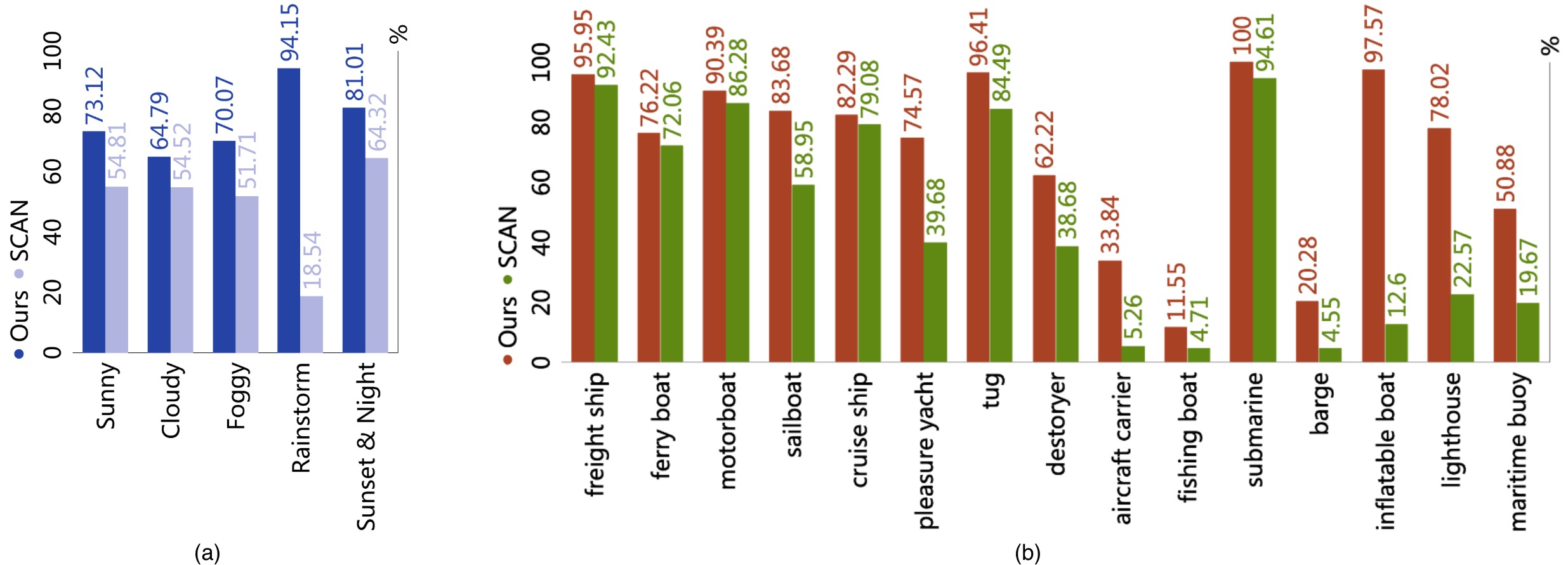}

  \caption{Validation experiments with our proposed method and SCAN. (a): Comparison of accuracy on RMO with multiple weather \& illumination conditions. (b): Comparison of accuracy on RMO with different categories.
  }
  \label{fig:category}
% \vspace{-1em}
\end{figure}

In Fig.\ref{fig:category}, we compare our proposed method with SCAN \cite{van2020scan}, an unsupervised clustering method that works only on unlabeled RMO.
Real-world maritime object images are rarely taken under bad weather or uneven illumination conditions, and have obvious long-tail issues. 
Therefore, only using real-world data for training easily leads to unbalanced classification results.

\textbf{\textit{Comparison of accuracy on RMO with multiple weather \& illumination conditions.}}
As shown in Fig. \ref{fig:category}(a), the UDA method we proposed using AIMO has achieved better performance under various weather and illumination conditions. 
Particularly in \textbf{\textit{rainstorm}} weather, our method has witnessed a significant performance boost. 
Moreover, in \textit{\textbf{foggy}} weather with insufficient lighting and potential object occlusion, the classification accuracy of our method is over 18 \% higher than that of the unsupervised classification method solely on RMO. Additionally, the introduction of AIMO has also improved the classification accuracy of maritime objects in other scenarios with relatively abundant lighting.
The results indicates that the application of generated data containing richer environmental information enhances the classification performance under various situations, which prove that the effectiveness of AIMO with multiple weather and time conditions.

\begin{table}[t]
    \caption{Image quality assessment on AIMO}
    \label{tab:IQA}
    \begin{tabular}{lccccc}
        \toprule
        Dataset  & ENTROPY  & NIQE  &  FID   &  CLIPIQA  \\
        \midrule
        AIMO  &  7.325 & 4.874 & 	34.011		&  0.784  \\
        \bottomrule 
    \end{tabular}
\end{table}

\begin{table}[t]
    \caption{Comparison with other UDA methods on RMO}
    \label{tab:comparison}
    \begin{tabular}{lc}
        \toprule
        Method  & Acc\ \%  \\
        \midrule
        ViT-small \cite{DBLP:conf/iclr/DosovitskiyB0WZ21} &  56.456   \\
        ViT-base  \cite{DBLP:conf/iclr/DosovitskiyB0WZ21}
        &  60.693  \\
        Swin-B \cite{liu2021swin}    & 58.761     \\
        CDTrans \cite{xu2021cdtrans} & 59.183 \\
        PMTrans \cite{zhu2023patch} & 65.463 \\
        TVT \cite{yang2023tvt}      & 62.260 \\
        % ConMix \cite{kumar2023conmix} & 59.737 \\
        PDA \cite{bai2024prompt}      & 65.891 \\ 
        UniMoS \cite{li2024split}    &  69.798 \\
        DAMP \cite{du2024domain}   &   72.398 \\
        \textbf{Ours}  &  \textbf{73.675} \\
        \bottomrule 
    \end{tabular}
\end{table}

\begin{table}[t]
    \caption{Comparison with other UDA methods on Mini-DomainNet}
    \label{tab:domainnet}
    \begin{tabular}{lc}
        \toprule
        Method  & Acc\ \%  \\
        \midrule
        ViT-base  \cite{DBLP:conf/iclr/DosovitskiyB0WZ21}
        &  65.7  \\
        CLIP \cite{radford2021learning} &   82.8 \\
        DAPromp \cite{DBLP:journals/tnn/GeHXLSLH25} &
        85.8 \\
        ADCLIP \cite{DBLP:conf/iccvw/SinghaPJB23} & 
        86.9 \\
        UniMoS \cite{li2024split}    &  87.3 \\
        DAMP \cite{du2024domain}   &   \textbf{87.6} \\
        \textbf{Ours}  &  87.5 \\
        \bottomrule 
    \end{tabular}
\end{table}

\textbf{\textit{Comparison of accuracy on RMO with different categories.}}
In Fig. \ref{fig:category}(b), the UDA method we proposed using AIMO has achieved better performance across all categories. In the categories of \textit{\textbf{pleasure yacht}}, \textit{\textbf{destroyer}}, \textbf{\textit{inflatable boat}}, \textit{\textbf{lighthouse}} and \textbf{\textit{maritime buoy}}, our method significantly improved the classification accuracy. 
We obverse that especially in the categories of \textbf{\textit{inflatable boat}}, \textit{\textbf{lighthouse}} and \textbf{\textit{maritime buoy}}, the classification accuracy of our method was notably different from that of the unsupervised clustering method, because these three categories account for a relatively small proportion in RMO, as shown in Fig. \ref{fig:data}(e). 
% However, for the \textit{\textbf{submarine}} and \textit{\textbf{barge}} categories in RMO which also have a small amount of data, whether using AIMO has a extremely minor impact on the results. 
% This maybe because the samples within \textit{\textbf{submarine}} category show high similarity, making it easier to recognize; while the barge category includes various types, and the diversity covered by AIMO is insufficient to cover the actual diversity.
% Similarly, the \textbf{\textit{aircraft carrier}} and \textbf{\textit{fishing boat}} categories have a strong diversity of objects, and our AIMO has limitations in these categories.
It is proved that by using the category balanced AIMO for transfer learning, we effectively improve the classification accuracy of categories with rare samples and alleviate the long-tail effect to a certain extent.

\subsection{Effectiveness of Domain Adaptation}
\label{sec:4.4}
We compare our proposed method with other UDA methods, e.g., ViT-small \cite{DBLP:conf/iclr/DosovitskiyB0WZ21}, ViT-base \cite{DBLP:conf/iclr/DosovitskiyB0WZ21}, Swin-B \cite{liu2021swin}, CDTrans \cite{xu2021cdtrans}, PMTrans \cite{zhu2023patch}, TVT \cite{yang2023tvt}, PDA \cite{bai2024prompt}, UniMoS \cite{li2024split}, and DAMP \cite{du2024domain} in Table \ref{tab:comparison}.
In summary, our proposed method performs better than other recent methods, and obtains an average classification accuracy of 73.675\% on RMO.

\textbf{\textit{Comparison with the basic networks.}}
ViT-small \cite{DBLP:conf/iclr/DosovitskiyB0WZ21} and ViT-base \cite{DBLP:conf/iclr/DosovitskiyB0WZ21} have applied Transformer to image classification for the first time, demonstrating the potential of global attention in visual tasks. 
On this basis, Swin Transformer \cite{liu2021swin} takes into account global modeling and local concentration bias through windowing and hierarchical design. 
Compared with these basic networks, our proposed method gained of over 10 \% on the classification accuracy, which attributes to our effective designs based on a series of modules in order to mitigate domain divergence.

\textbf{\textit{Comparison with methods using pseudo labels.}}
CDTrans \cite{xu2021cdtrans} learned domain-invariant features through cross-domain attention and used target domain pseudo labels for self-supervised learning.
Given the substantial differences in distribution between AIMO and RMO datasets, directly aligning image pairs from the source and target domains is not an effective approach. 
And the large domain gap makes the pseudo labels more unreliable, which directly affects the model performance.
Therefore, CDTrans \cite{xu2021cdtrans} exhibits suboptimal performance.
%Compared with CDTrans \cite{xu2021cdtrans}, we depend on adversarial training without pseudo labels.
Also, different from PMTrans \cite{zhu2023patch}, which constructed intermediate domains to alleviate domain gap and also relied on the pseudo labels during training, our method utilizes VLMs to alleviate the domain-shift from source domain to target domain, achieving nearly 8.2 \% gains on the accuracy.

\textbf{\textit{Comparison with method using adversarial adaptation.}}
TVT \cite{yang2023tvt} designed a transferability adaption module based on adversarial adaptation.
The learned transferabilities were injected into attention blocks, promoting ViT to focus on both domain-invariant and domain-specific features.  
Compared with TVT \cite{yang2023tvt}, which also utilizes adversarial training for UDA, the accuracy of our proposed method is 11 \% higher, probably due to the use of VLMs (i.e., CLIP) for improving robustness of the source domain features.

\textbf{\textit{Comparison with methods using VLMs.}}
% PDA \cite{bai2024prompt} performed a two-branch prompt tuning. 
% Firstly, the zero-shot capability of CLIP is used to make pseudo labels of target domain, and further extracted the text embeddings and image embeddings of source domain and target domain. 
% And then they used image embedding to guide the backbone network to learn the self-enhanced features and cross-domain features.
PDA \cite{bai2024prompt} utilized the zero-shot capability of CLIP to make pseudo labels of target domain, and further extracted the text embeddings and image embeddings of source domain and target domain. 
UniMoS \cite{li2024split} was a Unified Modality Separation framework for unsupervised domain adaptation, which distinctly disentangles CLIP's features into language-associated and vision-associated components.
DAMP \cite{du2024domain} mutually aligned visual and textual embeddings based on a cross-attention module, a semantic-consistency loss and an instance-discrimination contrastive loss.
Different from the methods above, we achieved better classification results which benefits from injecting the semantic information into backbone network through self-knowledge distillation loss.
Also, we utilize curriculum learning strategy to reduce data overfitting, which further improves performance.
% Compared with PDA \cite{bai2024prompt}, which also utilizes VLMs, our proposed method performs more than 7.78 \% on the classification accuracy.

\textbf{\textit{Comparison with methods on Mini-DomainNet.}}
To demonstrate the generality of our proposed method in UDA, we conduct comparative experiments with other methods on the Mini-DomainNet dataset, which is a subset of the most challenging dataset DomainNet \cite{DBLP:conf/iccv/PengBXHSW19} in Table \ref{tab:domainnet}.
The results proved the generalization of our proposed method.

% \begin{table}[t]
%     \caption{Ablation study for classification accuracy on RMO}
%     \label{tab:ablation}
%     \begin{tabular}{lllllc}
%         \toprule
%         \thead{Adversarial\\Training} & 
%         \thead{Input\\Offset} & 
%         \thead{Token\\Offset} & 
%         \thead{Generalization\\Enhancement} & 
%         \thead{Adaptive\\Adjustment} & 
%         \thead{Acc (\%)} \\
%         \midrule
%         \checkmark  &   &  &  &  & 61.614  \\
%         \checkmark  & \checkmark &  &  &  & 67.253 \\
%         \checkmark  &   & \checkmark &  &  & 69.385 \\
%         \checkmark  &   & \checkmark & \checkmark &  & 72.287 \\
%         \checkmark  &   & \checkmark & \checkmark & \checkmark & \textbf{73.315} \\ 
%         \bottomrule
%     \end{tabular}
% \end{table}

\begin{table}[t]
    \caption{Ablation study for classification accuracy on RMO}
    \label{tab:ablation}
    \begin{tabular}{lllllc}
        \toprule
        Adversarial Training & 
        Input Offset & 
        Token Offset & 
        Generalization Enhancement & 
        Adaptive Adjustment & 
        Acc (\%) \\
        \midrule
        \checkmark  &   &  &  &  & 61.629  \\
        \checkmark  & \checkmark &  &  &  & 67.416 \\
        \checkmark  &   & \checkmark &  &  & 69.443 \\
        \checkmark  &   & \checkmark & \checkmark &  & 71.337 \\
        \checkmark  &   & \checkmark & \checkmark & \checkmark & \textbf{73.675} \\ 
        \bottomrule
    \end{tabular}
\end{table}

\begin{table}[t]
    \caption{Classification accuracy on RMO with different prompt template}
    \label{tab:prompt}
    \begin{tabular}{lc}
        \toprule
        Prompt Template & Acc (\%) \\
        \midrule
        A photo in \{domain\} & 69.570 \\
        A photo of a \{class\} & 71.979 \\
        A photo of a \{class\} in \{domain\} & \textbf{73.675} \\ 
        \bottomrule
    \end{tabular}
\end{table}

\subsection{Ablation Study}
\label{sec:4.5}
For better understanding of the effect of each proposed module, we conduct an ablation study on RMO and show the result in Table \ref{tab:ablation}.

We combine ViT-B/16 \cite{DBLP:conf/iclr/DosovitskiyB0WZ21} with adversarial training as the baseline and obtain a 61.629\% on the classification accuracy of RMO.
Adding offsets on input images directly makes a 5.787\% increase compared to baseline in classification accuracy.
Comparison of the second and third rows in Table \ref{tab:ablation}, replacing the direct offset on images with adding random offset to the token sequence further improves the classification accuracy by 2.027\%.
It proves that injecting perturbations into the hidden space is a better choice.
On this basis, the \textit{Generalization Enhancement Module} is introduced to make the classification accuracy finally rise to 71.337\%, which shows that it is meaningful to use CLIP for Self-Knowledge Distillation on the source domain.
Finally, we employ curriculum learning to regularly select the source domain samples for training, which further improves the classification accuracy to 73.675 \%.
The result demonstrates the effectiveness of \textit{Adaptive Adjustment Mechanism}.
In summary, the ablation studies conducted on the RMO effectively validate the contributions of the proposed modules.
% In addition, we concretely show in Fig. \ref{fig:test} how the \textit{Adaptive Adjustment Mechanism} prevents the model collapse.
% We use the number of unique model predicted labels on the target domain to characterize the diversity of model predictions.
% During training, we detected a decrease in diversity twice, so the adaptation parameter $\mu$ was reset twice.
% Therefore, the classification accuracy steadily rises with the number of iterations, and there is no case of model collapse.

Moreover, we further compare the selection of prompt for text embedddings.
As shown in Table \ref{tab:prompt}, when we use \textit{"A photo of a \{class\} in \{domain\}"} as prompt template, the classification accuracy achieved the highest value.
This is due to the comprehensive consideration of the influence of category and weather on the classification accuracy. 
Compared with using \textit{"\{class\}"} or \textit{"\{domain\}"} alone, this compositive prompt template allows the generalization enhancement module to contain more information about the image.

\begin{figure}
% \vspace{-1em}
  \centering
  \includegraphics[width=0.8\linewidth]{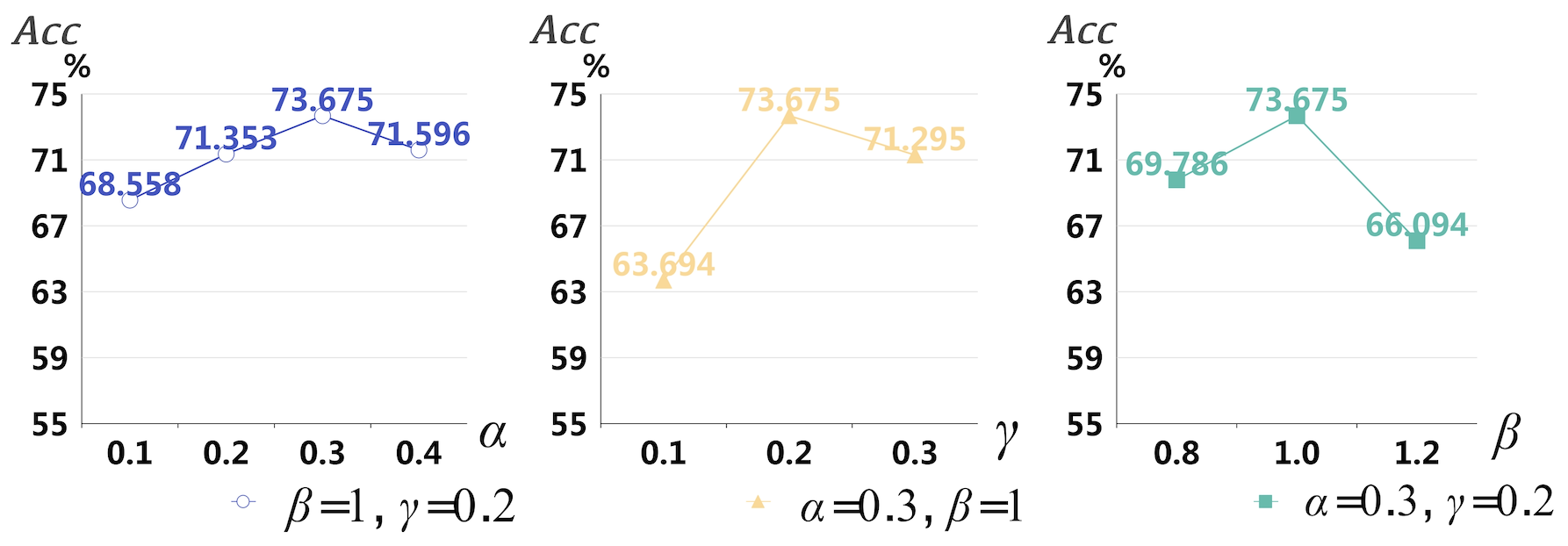}

  \caption{Parameter sensitivity experiments.
  }
  \label{fig:parameter}

\end{figure}

\subsection{Parameter Sensitivity Discussion}
\label{sec:4.6}
In this part, we analyze the sensitivity of three key parameters in our framework: $\alpha$ and $\beta$ used in Eq. \ref{loss_all}, and $\gamma$ in Eq. \ref{random offset}.
The classification accuracy trends under varying parameters are visualized in Fig. \ref{fig:parameter}.
% Fig. \ref{fig:parameter} (a), (b), and (c) are referred to $\alpha$, $\beta$, and $\gamma$, respectively.
As shown in Fig. \ref{fig:parameter}(a), the model maintains stable performance across $\alpha \in [0.1, 0.4]$, with accuracy fluctuating narrowly between 68.56\% and 73.68\%. 
This minimal variation validates the robustness of the proposed \textit{Generalization Enhancement} module.
We obverse that the choice of $\alpha$ has little impact on the performance of the model, which proves the stability of our proposed \textit{Generalization Enhancement Module}.
Fig. \ref{fig:parameter}(b) reveals a critical dependency on $\gamma$, because $\gamma$ influences the strength of regularization. 
When $\gamma=0.2$, the model achieves peak accuracy, but performance plummets a 9.98\% degradation at $\gamma=0.1$. 
This sharp decline underscores the necessity of sufficient regularization that overly small $\gamma$ fails to mitigate overfitting, while larger values retain competitive accuracy.
While, when the value of $\gamma$ is too small, the classification accuracy drops significantly, because the regularization effect is not achieved.
Similarly, the weight of the refinement loss, $\beta$, also has an impact on the classification accuracy of the model.
The classifier refinement loss weight $\beta$ exhibits a clear optimum at $\beta=1$. 
Deviations from this value, degrade performance by 3.9–7.6\%, as shown in Fig. \ref{fig:parameter}(c).
This implies that insufficient weight weaken its correction effect, while overemphasis may distort the primary learning objective.
Therefore, the best choices for these parameters are $\alpha=0.3$, $\beta=1$, and $\gamma=0.2$.
% In this part, we analyze the sensitivity of three key parameters in our framework: $\alpha$ and $\beta$ used in Eq. \ref{loss_all}, and $\gamma$ in Eq. \ref{random offset}.
% The classification accuracy trends under varying parameters are visualized in Fig. \ref{fig:parameter}.
% As shown in Fig. \ref{fig:parameter}(a), the stable variation validates the robustness of the proposed \textit{Generalization Enhancement} module.
% Fig. \ref{fig:parameter}(b) reveals a critical dependency on $\gamma$, which underscores that the necessity of sufficient regularization that overly small $\gamma$ fails to mitigate overfitting, while larger values retain competitive accuracy.
% And the results in Fig. \ref{fig:parameter}(c) imply that insufficient weight weaken its correction effect, while overemphasis may distort the primary learning objective.

% \begin{table}
%     \centering
%     \begin{tabular}{lr}
%         \toprule
%         $\alpha$  & Acc\ \%  \\
%        \midrule
%         0.1      & 71.221 \\
%         0.2      & \textbf{72.346} \\
%         0.3      & 69.697  \\
%         \bottomrule
%     \end{tabular}
%     \caption{Classification Accuracy on RMO with different
%     $\alpha$}
%     \label{tab:plain}
% \end{table}

\section{Conclusion}
In this work, we propose a label-rich and balanced AI-generated Benchmark for Maritime Object Classification (AIMO) with multiple weather and illumination conditions. 
And we construct a Real Maritime Object Benchmark (RMO) collected from a series of real maritime object datasets and the Google.
We further introduce an UDA framework with adversarial training and curriculum learning based on Vision Transformer to address maritime object classification in the real-world scenarios.
It leverages the generalization of VLM to refine source domain features and narrow domain gap. 
And an adaptive adjustment mechanism is applied to prevent model collapse and data overfitting.
Experiments validate the effectiveness of the proposed benchmark and unsupervised domain adaptation framework.

% \section*{Acknowledgments}
% This work was supported by the National Key Research and Development Program of China (2021YFF0704000) and the National Natural Science Foundation of China (U22A2068, U21B2024).

\printcredits

\bibliographystyle{cas-model2-names}

\bibliography{cas-refs}

@String(ICCV= {Int. Conf. Comput. Vis.})

@String(ICME = {Int. Conf. Multimedia and Expo})

@String(ICLR = {Int. Conf. Learn. Represent.})

@String(AAAI = {AAAI})

@String(ICCV  = {ICCV})

@String(ICME  =	{ICME})

@String(ICLR  = {ICLR})

@inproceedings{salem2022transfer,
  title={Transfer learning on efficientnet for maritime visible image classification},
  author={Salem, Mostafa Hamdy and Li, Yujian and Liu, Zhaoying},
  booktitle={2022 7th international conference on signal and image processing (ICSIP)},
  pages={514--520},
  year={2022},
  organization={IEEE}
}

@article{wang2020generalizing,
  title={Generalizing from a few examples: A survey on few-shot learning},
  author={Wang, Yaqing and Yao, Quanming and Kwok, James T and Ni, Lionel M},
  journal={ACM computing surveys (csur)},
  volume={53},
  number={3},
  pages={1--34},
  year={2020},
  publisher={ACM New York, NY, USA}
}

@article{zhao2020maritime,
  title={Maritime visible image classification based on double transfer method},
  author={Zhao, Ruiyu and Wang, Jianhua and Zheng, Xiang and Wen, Jing and Rao, Liuzhong and Zhao, Junxia},
  journal={IEEE Access},
  volume={8},
  pages={166335--166346},
  year={2020},
  publisher={IEEE}
}

@inproceedings{gundogdu2017marvel,
  title={Marvel: A large-scale image dataset for maritime vessels},
  author={Gundogdu, Erhan and Solmaz, Berkan and Y{\"u}cesoy, Veysel and Koc, Aykut},
  booktitle={Computer Vision--ACCV 2016: 13th Asian Conference on Computer Vision, Taipei, Taiwan, November 20-24, 2016, Revised Selected Papers, Part V 13},
  pages={165--180},
  year={2017},
  organization={Springer}
}

@article{iancu2021aboships,
  title={Aboships—an inshore and offshore maritime vessel detection dataset with precise annotations},
  author={Iancu, Bogdan and Soloviev, Valentin and Zelioli, Luca and Lilius, Johan},
  journal={Remote Sensing},
  volume={13},
  number={5},
  pages={988},
  year={2021},
  publisher={MDPI}
}

@inproceedings{zheng2020mcships,
  title={Mcships: A large-scale ship dataset for detection and fine-grained categorization in the wild},
  author={Zheng, Yitong and Zhang, Shun},
  booktitle={2020 IEEE International Conference on Multimedia and Expo (ICME)},
  pages={1--6},
  year={2020},
  organization={IEEE}
}

@article{meng2022fine,
  title={Fine-grained ship recognition for complex background based on global to local and progressive learning},
  author={Meng, Hao and Tian, Yang and Ling, Yue and Li, Tao},
  journal={IEEE Geoscience and Remote Sensing Letters},
  volume={19},
  pages={1--5},
  year={2022},
  publisher={IEEE}
}

@inproceedings{bloisi2015argos,
  title={ARGOS-Venice boat classification},
  author={Bloisi, Domenico D and Iocchi, Luca and Pennisi, Andrea and Tombolini, Luigi},
  booktitle={2015 12th IEEE International Conference on Advanced Video and Signal Based Surveillance (AVSS)},
  pages={1--6},
  year={2015},
  organization={IEEE}
}

@article{zhang2024viewpoint,
  title={A Viewpoint Adaptation Ensemble Contrastive Learning framework for vessel type recognition with limited data},
  author={Zhang, Xiaocai and Xiao, Zhe and Fu, Xiuju and Wei, Xiaoyang and Liu, Tao and Yan, Ran and Qin, Zheng and Zhang, Jianjia},
  journal={Expert Systems with Applications},
  volume={238},
  pages={122191},
  year={2024},
  publisher={Elsevier}
}

@article{tian2024fregnet,
  title={FREGNet: Ship Recognition Based on Feature Representation Enhancement and GCN Combiner in Complex Environment},
  author={Tian, Yang and Meng, Hao and Yuan, Fei},
  journal={IEEE Transactions on Intelligent Transportation Systems},
  year={2024},
  publisher={IEEE}
}

@article{liu2022dual,
  title={Dual-Channel and Two-Stage Dehazing Network for Promoting Ship Detection in Visual Perception System},
  author={Liu, Ting and Zhou, Baijun},
  journal={Mathematical Problems in Engineering},
  volume={2022},
  number={1},
  pages={8998743},
  year={2022},
  publisher={Wiley Online Library}
}

@article{sun2022irdclnet,
  title={IRDCLNet: Instance segmentation of ship images based on interference reduction and dynamic contour learning in foggy scenes},
  author={Sun, Yuxin and Su, Li and Luo, Yongkang and Meng, Hao and Zhang, Zhi and Zhang, Wen and Yuan, Shouzheng},
  journal={IEEE Transactions on Circuits and Systems for Video Technology},
  volume={32},
  number={9},
  pages={6029--6043},
  year={2022},
  publisher={IEEE}
}

@inproceedings{raza2022simuships,
  title={SimuShips-A High Resolution Simulation Dataset for Ship Detection with Precise Annotations},
  author={Raza, Minahil and Prokopova, Hanna and Huseynzade, Samir and Azimi, Sepinoud and Lafond, Sebastien},
  booktitle={OCEANS 2022, Hampton Roads},
  pages={1--5},
  year={2022},
  organization={IEEE}
}

@misc{
you2024are,
title={Are Images Indistinguishable to Humans Also Indistinguishable to Classifiers?},
author={Zebin You and Xinyu Zhang and Hanzhong Allan Guo and Jingdong Wang and Chongxuan Li},
year={2024},
url={https://openreview.net/forum?id=MRnZ1KEXSt}
}

@inproceedings{radford2021learning,
  title={Learning transferable visual models from natural language supervision},
  author={Radford, Alec and Kim, Jong Wook and Hallacy, Chris and Ramesh, Aditya and Goh, Gabriel and Agarwal, Sandhini and Sastry, Girish and Askell, Amanda and Mishkin, Pamela and Clark, Jack and others},
  booktitle={International conference on machine learning},
  pages={8748--8763},
  year={2021},
  organization={PMLR}
}

@inproceedings{bai2024prompt,
  title={Prompt-based distribution alignment for unsupervised domain adaptation},
  author={Bai, Shuanghao and Zhang, Min and Zhou, Wanqi and Huang, Siteng and Luan, Zhirong and Wang, Donglin and Chen, Badong},
  booktitle={Proceedings of the AAAI Conference on Artificial Intelligence},
  volume={38},
  number={2},
  pages={729--737},
  year={2024}
}

@inproceedings{sun2022safe,
  title={Safe self-refinement for transformer-based domain adaptation},
  author={Sun, Tao and Lu, Cheng and Zhang, Tianshuo and Ling, Haibin},
  booktitle={Proceedings of the IEEE/CVF conference on computer vision and pattern recognition},
  pages={7191--7200},
  year={2022}
}

@inproceedings{DBLP:conf/iclr/DosovitskiyB0WZ21,
  author       = {Alexey Dosovitskiy and
                  Lucas Beyer and
                  Alexander Kolesnikov and
                  Dirk Weissenborn and
                  Xiaohua Zhai and
                  Thomas Unterthiner and
                  Mostafa Dehghani and
                  Matthias Minderer and
                  Georg Heigold and
                  Sylvain Gelly and
                  Jakob Uszkoreit and
                  Neil Houlsby},
  title        = {An Image is Worth 16x16 Words: Transformers for Image Recognition
                  at Scale},
  booktitle    = {9th International Conference on Learning Representations, {ICLR} 2021,
                  Virtual Event, Austria, May 3-7, 2021},
  publisher    = {OpenReview.net},
  year         = {2021},
  url          = {https://openreview.net/forum?id=YicbFdNTTy},
  bibsource    = {dblp computer science bibliography, https://dblp.org}
}

@inproceedings{ganin2015unsupervised,
  title={Unsupervised domain adaptation by backpropagation},
  author={Ganin, Yaroslav and Lempitsky, Victor},
  booktitle={International conference on machine learning},
  pages={1180--1189},
  year={2015},
  organization={PMLR}
}

@inproceedings{addepalli2024leveraging,
  title={Leveraging Vision-Language Models for Improving Domain Generalization in Image Classification},
  author={Addepalli, Sravanti and Asokan, Ashish Ramayee and Sharma, Lakshay and Babu, R Venkatesh},
  booktitle={Proceedings of the IEEE/CVF Conference on Computer Vision and Pattern Recognition},
  pages={23922--23932},
  year={2024}
}

@inproceedings{lai2024empowering,
  title={Empowering unsupervised domain adaptation with large-scale pre-trained vision-language models},
  author={Lai, Zhengfeng and Bai, Haoping and Zhang, Haotian and Du, Xianzhi and Shan, Jiulong and Yang, Yinfei and Chuah, Chen-Nee and Cao, Meng},
  booktitle={Proceedings of the IEEE/CVF Winter Conference on Applications of Computer Vision},
  pages={2691--2701},
  year={2024}
}

@inproceedings{verma2019manifold,
  title={Manifold mixup: Better representations by interpolating hidden states},
  author={Verma, Vikas and Lamb, Alex and Beckham, Christopher and Najafi, Amir and Mitliagkas, Ioannis and Lopez-Paz, David and Bengio, Yoshua},
  booktitle={International conference on machine learning},
  pages={6438--6447},
  year={2019},
  organization={PMLR}
}

@misc{pyiqa,
  title={{IQA-PyTorch}: PyTorch Toolbox for Image Quality Assessment},
  author={Chaofeng Chen and Jiadi Mo},
  year={2022},
  howpublished = "[Online]. Available: \url{https://github.com/chaofengc/IQA-PyTorch}"
}

@inproceedings{liu2021swin,
  title={Swin transformer: Hierarchical vision transformer using shifted windows},
  author={Liu, Ze and Lin, Yutong and Cao, Yue and Hu, Han and Wei, Yixuan and Zhang, Zheng and Lin, Stephen and Guo, Baining},
  booktitle={Proceedings of the IEEE/CVF international conference on computer vision},
  pages={10012--10022},
  year={2021}
}

@article{xu2021cdtrans,
  title={Cdtrans: Cross-domain transformer for unsupervised domain adaptation},
  author={Xu, Tongkun and Chen, Weihua and Wang, Pichao and Wang, Fan and Li, Hao and Jin, Rong},
  journal={arXiv preprint arXiv:2109.06165},
  year={2021}
}

@inproceedings{yang2023tvt,
  title={Tvt: Transferable vision transformer for unsupervised domain adaptation},
  author={Yang, Jinyu and Liu, Jingjing and Xu, Ning and Huang, Junzhou},
  booktitle={Proceedings of the IEEE/CVF Winter Conference on Applications of Computer Vision},
  pages={520--530},
  year={2023}
}

@inproceedings{zhu2023patch,
  title={Patch-mix transformer for unsupervised domain adaptation: A game perspective},
  author={Zhu, Jinjing and Bai, Haotian and Wang, Lin},
  booktitle={Proceedings of the IEEE/CVF conference on computer vision and pattern recognition},
  pages={3561--3571},
  year={2023}
}

@inproceedings{van2020scan,
  title={Scan: Learning to classify images without labels},
  author={Van Gansbeke, Wouter and Vandenhende, Simon and Georgoulis, Stamatios and Proesmans, Marc and Van Gool, Luc},
  booktitle={European conference on computer vision},
  pages={268--285},
  year={2020},
  organization={Springer}
}

@misc{zewe2022machine,
  title={{\guillemotright} In machine learning, synthetic data can offer real performance improvements {\guillemotleft}},
  author={Zewe, Adam},
  year={2022},
  publisher={MIT News Office}
}

@inproceedings{DBLP:conf/iclr/HeS0XZTBQ23,
  author       = {Ruifei He and
                  Shuyang Sun and
                  Xin Yu and
                  Chuhui Xue and
                  Wenqing Zhang and
                  Philip H. S. Torr and
                  Song Bai and
                  Xiaojuan Qi},
  title        = {Is Synthetic Data from Generative Models Ready for Image Recognition?},
  booktitle    = {The Eleventh International Conference on Learning Representations,
                  {ICLR} 2023, Kigali, Rwanda, May 1-5, 2023},
  publisher    = {OpenReview.net},
  year         = {2023},
  url          = {https://openreview.net/forum?id=nUmCcZ5RKF},
  bibsource    = {dblp computer science bibliography, https://dblp.org}
}

@article{shumailov2024ai,
  title={AI models collapse when trained on recursively generated data},
  author={Shumailov, Ilia and Shumaylov, Zakhar and Zhao, Yiren and Papernot, Nicolas and Anderson, Ross and Gal, Yarin},
  journal={Nature},
  volume={631},
  number={8022},
  pages={755--759},
  year={2024},
  publisher={Nature Publishing Group UK London}
}

@article{zhou2021image,
  title={An image-based benchmark dataset and a novel object detector for water surface object detection},
  author={Zhou, Zhiguo and Sun, Jiaen and Yu, Jiabao and Liu, Kaiyuan and Duan, Junwei and Chen, Long and Chen, CL Philip},
  journal={Frontiers in Neurorobotics},
  volume={15},
  pages={723336},
  year={2021},
  publisher={Frontiers Media SA}
}

@article{russakovsky2015imagenet,
  title={Imagenet large scale visual recognition challenge},
  author={Russakovsky, Olga and Deng, Jia and Su, Hao and Krause, Jonathan and Satheesh, Sanjeev and Ma, Sean and Huang, Zhiheng and Karpathy, Andrej and Khosla, Aditya and Bernstein, Michael and others},
  journal={International journal of computer vision},
  volume={115},
  pages={211--252},
  year={2015},
  publisher={Springer}
}

@inproceedings{le2011optimization,
  title={On optimization methods for deep learning},
  author={Le, Quoc V and Ngiam, Jiquan and Coates, Adam and Lahiri, Abhik and Prochnow, Bobby and Ng, Andrew Y},
  booktitle={Proceedings of the 28th international conference on international conference on machine learning},
  pages={265--272},
  year={2011}
}

@inproceedings{liu2022swin,
  title={Swin transformer v2: Scaling up capacity and resolution},
  author={Liu, Ze and Hu, Han and Lin, Yutong and Yao, Zhuliang and Xie, Zhenda and Wei, Yixuan and Ning, Jia and Cao, Yue and Zhang, Zheng and Dong, Li and others},
  booktitle={Proceedings of the IEEE/CVF conference on computer vision and pattern recognition},
  pages={12009--12019},
  year={2022}
}

@inproceedings{kang2019contrastive,
  title={Contrastive adaptation network for unsupervised domain adaptation},
  author={Kang, Guoliang and Jiang, Lu and Yang, Yi and Hauptmann, Alexander G},
  booktitle={Proceedings of the IEEE/CVF conference on computer vision and pattern recognition},
  pages={4893--4902},
  year={2019}
}

@article{zellinger2017central,
  title={Central moment discrepancy (cmd) for domain-invariant representation learning},
  author={Zellinger, Werner and Grubinger, Thomas and Lughofer, Edwin and Natschl{\"a}ger, Thomas and Saminger-Platz, Susanne},
  journal={arXiv preprint arXiv:1702.08811},
  year={2017}
}

@article{li2020maximum,
  title={Maximum density divergence for domain adaptation},
  author={Li, Jingjing and Chen, Erpeng and Ding, Zhengming and Zhu, Lei and Lu, Ke and Shen, Heng Tao},
  journal={IEEE transactions on pattern analysis and machine intelligence},
  volume={43},
  number={11},
  pages={3918--3930},
  year={2020},
  publisher={IEEE}
}

@inproceedings{sun2016return,
  title={Return of frustratingly easy domain adaptation},
  author={Sun, Baochen and Feng, Jiashi and Saenko, Kate},
  booktitle={Proceedings of the AAAI conference on artificial intelligence},
  volume={30},
  number={1},
  year={2016}
}

@inproceedings{mei2020instance,
  title={Instance adaptive self-training for unsupervised domain adaptation},
  author={Mei, Ke and Zhu, Chuang and Zou, Jiaqi and Zhang, Shanghang},
  booktitle={Computer Vision--ECCV 2020: 16th European Conference, Glasgow, UK, August 23--28, 2020, Proceedings, Part XXVI 16},
  pages={415--430},
  year={2020},
  organization={Springer}
}

@inproceedings{zhai2022lit,
  title={Lit: Zero-shot transfer with locked-image text tuning},
  author={Zhai, Xiaohua and Wang, Xiao and Mustafa, Basil and Steiner, Andreas and Keysers, Daniel and Kolesnikov, Alexander and Beyer, Lucas},
  booktitle={Proceedings of the IEEE/CVF conference on computer vision and pattern recognition},
  pages={18123--18133},
  year={2022}
}

@inproceedings{du2024domain,
  title={Domain-agnostic mutual prompting for unsupervised domain adaptation},
  author={Du, Zhekai and Li, Xinyao and Li, Fengling and Lu, Ke and Zhu, Lei and Li, Jingjing},
  booktitle={Proceedings of the IEEE/CVF Conference on Computer Vision and Pattern Recognition},
  pages={23375--23384},
  year={2024}
}

@inproceedings{cheng2024disentangled,
  title={Disentangled prompt representation for domain generalization},
  author={Cheng, De and Xu, Zhipeng and Jiang, Xinyang and Wang, Nannan and Li, Dongsheng and Gao, Xinbo},
  booktitle={Proceedings of the IEEE/CVF Conference on Computer Vision and Pattern Recognition},
  pages={23595--23604},
  year={2024}
}

@article{DBLP:journals/corr/abs-2502-04628,
  author       = {Runqing Jiang and
                  Ye Zhang and
                  Longguang Wang and
                  Pengpeng Yu and
                  Yulan Guo},
  title        = {AIQViT: Architecture-Informed Post-Training Quantization for Vision
                  Transformers},
  journal      = {CoRR},
  volume       = {abs/2502.04628},
  year         = {2025},
  url          = {https://doi.org/10.48550/arXiv.2502.04628},
  doi          = {10.48550/ARXIV.2502.04628},
  eprinttype    = {arXiv},
  eprint       = {2502.04628},
  bibsource    = {dblp computer science bibliography, https://dblp.org}
}

@article{wang2021survey,
  title={A survey on curriculum learning},
  author={Wang, Xin and Chen, Yudong and Zhu, Wenwu},
  journal={IEEE transactions on pattern analysis and machine intelligence},
  volume={44},
  number={9},
  pages={4555--4576},
  year={2021},
  publisher={IEEE}
}

@article{ha2016hypernetworks,
  title={Hypernetworks},
  author={Ha, David and Dai, Andrew and Le, Quoc V},
  journal={arXiv preprint arXiv:1609.09106},
  year={2016}
}

@article{gal2022image,
  title={An image is worth one word: Personalizing text-to-image generation using textual inversion},
  author={Gal, Rinon and Alaluf, Yuval and Atzmon, Yuval and Patashnik, Or and Bermano, Amit H and Chechik, Gal and Cohen-Or, Daniel},
  journal={arXiv preprint arXiv:2208.01618},
  year={2022}
}

@inproceedings{li2024split,
  title={Split to merge: Unifying separated modalities for unsupervised domain adaptation},
  author={Li, Xinyao and Li, Yuke and Du, Zhekai and Li, Fengling and Lu, Ke and Li, Jingjing},
  booktitle={Proceedings of the IEEE/CVF Conference on Computer Vision and Pattern Recognition},
  pages={23364--23374},
  year={2024}
}

@article{DBLP:journals/ipm/ZhangZHM24,
  author       = {Bo Zhang and
                  Xiao{-}Ming Zhang and
                  Feiran Huang and
                  Dezhuang Miao},
  title        = {Cross-domain knowledge collaboration for blending-target domain adaptation},
  journal      = {Inf. Process. Manag.},
  volume       = {61},
  number       = {4},
  pages        = {103730},
  year         = {2024},
  url          = {https://doi.org/10.1016/j.ipm.2024.103730},
  doi          = {10.1016/J.IPM.2024.103730},
  bibsource    = {dblp computer science bibliography, https://dblp.org}
}

@article{zhan2026dual,
  title={Dual transferable knowledge interaction for source-free domain adaptation},
  author={Zhan, Mengmeng and Wu, Zongqian and Yang, Jiaying and Peng, Lin and Shen, Jialie and Zhu, Xiaofeng},
  journal={Information Processing \& Management},
  volume={63},
  number={1},
  pages={104302},
  year={2026},
  publisher={Elsevier}
}

@article{zhang2025moment,
  title={Moment matching of joint distributions for unsupervised domain adaptation},
  author={Zhang, Bo and Zhang, Xiaoming and Zhou, Zhibo and Liu, Yun and Li, Yancong and Huang, Feiran},
  journal={Information Processing \& Management},
  volume={62},
  number={1},
  pages={103944},
  year={2025},
  publisher={Elsevier}
}

@inproceedings{DBLP:conf/iccv/PengBXHSW19,
  author       = {Xingchao Peng and
                  Qinxun Bai and
                  Xide Xia and
                  Zijun Huang and
                  Kate Saenko and
                  Bo Wang},
  title        = {Moment Matching for Multi-Source Domain Adaptation},
  booktitle    = {2019 {IEEE/CVF} International Conference on Computer Vision, {ICCV}
                  2019, Seoul, Korea (South), October 27 - November 2, 2019},
  pages        = {1406--1415},
  publisher    = {{IEEE}},
  year         = {2019},
  url          = {https://doi.org/10.1109/ICCV.2019.00149},
  doi          = {10.1109/ICCV.2019.00149},
  bibsource    = {dblp computer science bibliography, https://dblp.org}
}

@article{DBLP:journals/tnn/GeHXLSLH25,
  author       = {Chunjiang Ge and
                  Rui Huang and
                  Mixue Xie and
                  Zihang Lai and
                  Shiji Song and
                  Shuang Li and
                  Gao Huang},
  title        = {Domain Adaptation via Prompt Learning},
  journal      = {{IEEE} Trans. Neural Networks Learn. Syst.},
  volume       = {36},
  number       = {1},
  pages        = {1160--1170},
  year         = {2025},
  url          = {https://doi.org/10.1109/TNNLS.2023.3327962},
  doi          = {10.1109/TNNLS.2023.3327962},
  bibsource    = {dblp computer science bibliography, https://dblp.org}
}

@inproceedings{DBLP:conf/iccvw/SinghaPJB23,
  author       = {Mainak Singha and
                  Harsh Pal and
                  Ankit Jha and
                  Biplab Banerjee},
  title        = {{AD-CLIP:} Adapting Domains in Prompt Space Using {CLIP}},
  booktitle    = {{IEEE/CVF} International Conference on Computer Vision, {ICCV} 2023
                  - Workshops, Paris, France, October 2-6, 2023},
  pages        = {4357--4366},
  publisher    = {{IEEE}},
  year         = {2023},
  url          = {https://doi.org/10.1109/ICCVW60793.2023.00470},
  doi          = {10.1109/ICCVW60793.2023.00470},
  bibsource    = {dblp computer science bibliography, https://dblp.org}
}

\end{document}